\documentclass[runningheads]{llncs}
\usepackage{url}
\usepackage{amsmath}
\usepackage[shortlabels]{enumitem}
\usepackage{booktabs} 
\usepackage{subcaption} 
\usepackage{graphicx}
\usepackage{pgfplots}
\usepackage{multirow}
\usepackage[all]{nowidow}
\usepackage[utf8]{inputenc}
\usepackage{tikz}
\usetikzlibrary{shapes.geometric}
\usepackage{multicol}
\usepackage{multirow}
\usepackage{algpseudocode,algorithm,algorithmicx}
\usepackage{caption}
\usepackage[export]{adjustbox}
\usepackage{amsmath}
\usepackage{amsfonts}
\usepackage[normalem]{ulem} 

\usepackage{enumitem} 
\definecolor{blue}{HTML}{1F77B4}
\definecolor{orange}{HTML}{FF7F0E}
\definecolor{green}{HTML}{2CA02C}

\pgfplotsset{compat=1.14}

\begin{document}
\title{On The Effects Of Data Normalisation For Domain Adaptation On EEG Data}
%
%
\author{Andrea Apicella\inst{1,2} \and
Francesco Isgrò\inst{1,2} \and Andrea Pollastro\inst{1,2} \and Roberto Prevete\inst{1,2}}
%
%
\institute{Department of Electrical Engineering and Information Technology, University of Naples Federico II, Naples, Italy
\and
Laboratory of Augmented Reality for Health Monitoring (ARHeMLab)}
%


\maketitle              
%
\begin{abstract}
\let\thefootnote\relax\footnotetext{This paper has been published in its final version on \textit{Engineering Applications of Artificial Intelligence} journal with DOI \url{https://doi.org/10.1016/j.engappai.2023.106205}} In the Machine Learning (ML) literature, a well-known problem is the \textit{Dataset Shift} problem where, differently from the ML standard hypothesis, the data in the
training and test sets can follow different probability distributions, leading ML systems toward poor generalisation performances. This problem is intensely felt in the Brain-Computer Interface (BCI) context, where bio-signals as Electroencephalographic (EEG) are often used. In fact, EEG signals are highly non-stationary both over time and between different subjects. 
To overcome this problem, several proposed solutions are based on recent transfer learning approaches such as Domain Adaption (DA). In several cases, however, the actual causes of the improvements remain ambiguous. This paper focuses on the impact of data normalisation, or standardisation strategies applied together with DA methods. In particular, using \textit{SEED}, \textit{DEAP}, and \textit{BCI Competition IV 2a} EEG datasets, we experimentally evaluated the impact of different normalization strategies applied with and without several well-known DA methods, comparing the obtained performances.
It results that the choice of the normalisation  strategy plays a key role on the classifier performances in DA scenarios, and interestingly, in several cases, the use of only an appropriate normalisation schema outperforms the DA technique.


\end{abstract}
\section{Introduction}
\label{sec:introduction}
In recent years, Brain-Computer Interfaces (BCIs) have been emerging as technology allowing the human brain to communicate with external devices without the use of peripheral nerves and muscles, enhancing the interaction capability of the user with the environment. BCI applications go from severely disabled persons for rehabilitation purposes to healthy subjects for devising new types of applications \cite{muhl2014survey}. In particular, BCI has a growing interest in the scientific community thanks to its implication in several medical fields, such as assisting  \cite{khan2020review}, monitoring \cite{arpaia2021design}, enhancing \cite{huang2014focus}, or diagnosing patients’ emotional or physical states \cite{arpaia2022successfully,apicella2021eeg}. 
Current literature reports that patients subjected to BCI-based Rehabilitation methods show benefit and improvement in their injured capacities \cite{gumuslu2020emotion}. 
Currently, several methods exist to allow the interaction between humans and machines. In particular, several proposals for BCI methods based on Electroencephalographic (EEG) signals are made. This is because measuring and monitoring the brain's electrical activity can provide important information related to the brain's physiological, functional, and pathological status. EEG signals are particularly suitable for this aim thanks to their essential qualities, such as non-invasiveness and high temporal resolution.  

Modern Machine Learning (ML) methods such as Deep Neural Networks (DNNs) are mainly used to process acquired EEG signals for several tasks, such as emotion classification, engagement and attention detection. In general, a supervised ML model learns from human classified data to generalise to new unknown data. The standard pipeline to develop an ML system consists in i) data acquisition, ii) data preprocessing, iii) feature extraction, iv) model learning v) model validation. However, the performance obtained using classical ML methods in EEG-related tasks is often poor \cite{li2018hierarchical}. This is mainly because the EEG signal is highly non-stationary \cite{huang2022generator}, substantial differences across the EEG acquired at different times or from different subjects exist, even with the same affect felt. More in detail, the starting hypothesis of the traditional ML methods states that all the used data, whether used in the training process or not, come from the same probability distribution. This assumption results are not always verified in the case of EEG signals. In the ML literature, this is an instance of the Dataset Shift problem \cite{quinonero2008dataset}. In a nutshell, a Dataset Shift arises when the starting ML assumption is not valid, so the distribution of the training data differs from the data distribution used outside of the training stage. In other words, a model trained on a set of EEG data acquired from a given subject at a specific time  (or during a specific session) should not work as expected in classifying EEG signals acquired from a different subject at different times. In other words, the model has poor generalisation performance. A first attempt to mitigate this problem is training specific models for each subject (Subject-Dependent models) to reduce the performance gap due to using the same ML system on different users. However, non-stationary signal problems related to the different user's physical and psychological conditions at different times remain. Furthermore, a Subject-Dependent model is valid only for the subject providing training data acquisition, making these models expensive and not very versatile and uncomfortable to the user, who will be tied to initial acquisition sessions before it can actually use the system for real classifications.

For these reasons, newer studies \cite{zhou2022domain,kouw2019review} tried to overcome these limits given by Dataset Shift, taking into account the difference between the data distribution probabilities (\textit{domains}) acquired in different times and for different subjects. Several proposed solutions are based on Transfer Learning (TL) \cite{pan2009survey}, a set of approaches aiming to transfer the knowledge learned from a system to improve another. TL approaches can be categorised into several subfamilies. One of the most famous is the Domain Adaptation (DA)\cite{kouw2019review} approaches family. DA approaches start from the hypothesis that unlabeled data from the target domain are also available during the training stage. For example, in the case of EEG-based emotion recognition, class-labelled data can be acquired in an initial session and classified using a standardised labelling protocol (e.g., questionnaires administered during the task). In contrast, class-unlabeled data can be acquired in a later session. DA provides several methods exploiting both labeled and unlabeled data to build an ML model able to minimise the discrepancy between the two data distributions, leading to better classification performances on unlabeled data.
Thus, performance improvements are often reported using DA methods in several EEG-based classification studies. However, from a methodological point of view, it is essential to note that the pipeline to develop and evaluate an ML model consists of several steps which can influence each other \cite{de2019understanding}. Consequently, in several cases \cite{ganin2015unsupervised} the causes of the improvements can remain ambiguous. This paper focuses on the impact of data normalisation, or standardisation strategies applied together with DA methods. 

However, DA methods assume that all the class-labelled data used during the training comes from the same source probability distribution (\textit{source domain}), i.e. all the labelled data belong to the same unique domain. This assumption is often neglected in several EEG-based works  \cite{zheng2015transfer,chai2017fast}, considering all the labeled data together during the training stage. Indeed, in several cross-subject/cross-session studies adopting DA strategies, it is not hard to see attempts to generalise toward an unseen domain (a subject or a session) using learning/source data acquisitions from several other and different sessions/subjects without considering their different probability distributions, so treated as belonging to the same domain. Despite this, performance improvements are often reported using DA methods in several EEG-based classification studies. We hypothesise that this improvement may not be caused by the DA method but by some data normalisation or standardisation strategies applied a priori. 

More in detail, in ML applications,  \textit{normalisation functions}\cite{singh2020investigating} are often applied to pre-process the input features before to be fed to the ML system. Normalisation functions are often adopted to scale or transform the features such that each feature has a uniform
contribution to the ML pipeline. In \cite{singh2020investigating} is shown that using some normalisation function can impact or not on the final classification performance, depending on the different features and properties that
data may have. However, several tasks involving EEG and ML methods applying well-known normalisation functions (such as Z-score normalisation\cite{singh2020investigating}) on the input features have been proposed over the years (for example, \cite{duan2021decoding}). In many of these studies, the normalisation function is often a de-facto standard in an EEG ML pipeline. In particular, one of the most used normalisation strategies is the Z-score normalisation, consisting of a translation and a scaling of the data with respect to its mean and variance. For instance, in \cite{chen2021personal,apicella2022eeg,fernandez2021cross,fan2017eeg,arevalillo2019combining} is shown that using a normalisation function can affect the cross-subject performances. In particular, the translation with respect to the mean can already be seen as a simple form of domain adaptation. 

This study aims to investigate if and how some normalisation strategies affect the performance of some of the DA methods applied to EEG signal classification. The main contribution of this research work is that in several EEG classification problems, the higher impact in reducing the domain shift seems to be due mainly to the data normalisation stage rather than the application of several DA methods commonly used in the literature.

The paper is organised as follows: in Section \ref{sec:related_works} some of the most known DA methods are reported, in Section \ref{sec:problem_description} the DA framework is described, and our hypothesis is expressed, in Section \ref{sec:experimental_assessment} the experimental assessment, and the obtained results are reported, in Section \ref{sec:discussion} the obtained results are discussed. Finally, Section \ref{sec:conclusions} is left to the final remarks.


\section{Related works}
\label{sec:related_works}
As in this work, we want to investigate the impact of input normalization strategies on DA methods. We first discuss DA approaches. Then, we present the main standard data normalization techniques in this context. Finally, we highlight differences and similarities with related research studies.

More recently, Transfer Learning (TL) methods are receiving strong attention from the scientific community. TL methods are based on the concept of Domain. Following the survey of Pan et al. \cite{pan2009survey}, a Domain can be defined as a set $D=\{F, P(X)\}$ where $F$ is a feature space and $P(X)$ is the marginal probability distribution of a specific dataset $X=\{x_1,x_2,\dots,x_n\} \in F$. Domain Adaptation methods start from the hypothesis that data sampled from two different Domains are available, called \textit{Source} Domain and \textit{Target} Domain, respectively. The main difference between Source and Target is that, while both data and labels $S_{Source}=\{(x_i, y_i)\}_{i=1}^n$ can be sampled from the Source domain, only feature data points $X_{Target}=\{x_j\}_{j=1}^m \in F_{Target}$ sampled from the Target Domain are available during the training stage, without any knowledge (unsupervised DA) or minimal knowledge (semi-supervised DA) of their real labels. DA methods are getting a great deal of attention in the scientific community in different contexts, such as image classification, voice recognition, etc., and several proposals have been made over the years. One trend of the literature is to adapt DA methods originally proposed in a context (e.g., image classification) to another one (e.g., EEG emotion recognition). For example, in \cite{hagad2021learning} methods to adapt DA strategies from the image classification context to EEG emotion classification are proposed. However, each context has its characteristics and peculiarities, making it not trivial to adapt a DA method from one task to another. The scientific community attempted to adapt well-established DA methods to tasks involving EEG signal processing in the emotion recognition field.  

In \cite{ganin2015unsupervised}, DA methods are divided into two main categories: i) shallow DA methods, where a representation function projecting the source and the target data is given a-priori, and deep DA methods, where the data representation is learned as part of the DA strategy.

For instance, one of the most known shallow DA methods is Transfer Component Analysis (TCA, \cite{pan2010domain}). TCA searches for a data transformation based on the Maximum Mean Discrepancy (MMD,\cite{gretton2006kernel}).
MMD was proposed to test the similarity between two probability distributions. An empirical estimation of MMD is given by 
$$MMD(X_S, X_T) = ||\frac{1}{|X_S|}\sum\limits_{i=1}^{|X_S|}\phi(\mathbf{x}^{(i)}_S) - \frac{1}{|X_T|}\sum\limits_{i=1}^{|X_T|}\phi(\mathbf{x}^{(i)}_T)||_H^2$$
where $X_S=\{\mathbf{x}^{(i)}_S\}_{i=1}^M$ and $X_T=\{\mathbf{x}^{(i)}_T\}_{i=1}^N$ are data sampled from the source and the target domain respectively, while $\phi(\cdot)$ is an appropriate feature mapping.

Starting from the hypothesis that the data are sampled from two different domains, TCA searches for a transformation of the data such that the data variance is maximally preserved reducing, at the same time, the MMD discrepancy between the domains distributions.

An evaluation of the TCA on EEG data for emotion recognition was made in \cite{zheng2015transfer}.
While it is not specifically proposed for Domain Adaptation, Kernel-PCA (KPCA,\cite{scholkopf1997kernel}) can be viewed as another shallow-DA strategy. In a nutshell, KPCA uses the kernel trick to project the data into proper kernel space and then apply the PCA to the projected data.

On another side, many modern deep DA strategies rely on Domain Adversarial Learning approaches, proposed in \cite{ganin2015unsupervised,ajakan2014domain,ganin2016domain}. In a nutshell, these proposals learn a DNN feature representation considering both the desired task and the discrepancy between the Source and the Target domain. The goal is to make the data distributions indistinguishable for an ad-hoc domain discriminator. The final model is a deep neural network model (Domain Adversarial Neural Network, DANN) predicting, for each input, both the corresponding class and the belonging domain. Therefore, learning a feature mapping that maximises the class prediction performances and the domain classification loss to make the feature distributions as similar as possible is made.
Adversarial Discriminative Domain Adaptation (ADDA) is another Domain adversarial learning strategy proposed in \cite{tzeng2017adversarial}. Differently from DANN,  ADDA learns two autoencoders $E_S$ and $E_T$, to represent the Source and the Target domains, respectively. Furthermore, $E_S$ is trained together with a classifier $C$, exploiting the available Source domain labelled data. Then, through an adversarial learning procedure, $E_T$ is trained to map the Target domain data to the space of the $E_S$ outputs. Finally, target data in $E_S$ can be classified by $C$.

Domain adversarial learning methods are widely used in several studies for EEG data recognition, for example, in \cite{tzeng2017adversarial,bao2020two,li2019regional}.

All the methods mentioned above only consider two domains: the Source and the Target one.

However, simple methods used to reduce gaps between different data relied on data normalisation schemes, such as min-max or $z$-score normalisation, where data are transformed using simple functions that leverage statistics on the data itself. For instance, in \cite{chen2021personal,apicella2022eeg,fernandez2021cross,fan2017eeg,arevalillo2019combining} is shown that just a proper normalisation schemes to preprocess the EEG data can affect the cross-subject performances.

In \cite{chen2021personal} several normalisation schemes were applied following two different schemas: i) All-subjects, where the whole dataset was normalised, ii) Single-subject, where the normalisation is made individually for each subject. The All-subject schema is the most common method used to mitigate the impact of each data value on the entire dataset. Single-subject, instead, consider each subject individually, applying normalisation to each subject. The authors empirically showed that Single-subject Z-score performs better in an EEG emotion recognition problem with respect to other normalisation schemes as min-max normalisation. 

In \cite{apicella2022eeg} single-subject Z-score normalisation is effectively used to improve the performance in the cross-subject case of a student engagement detection problem. \cite{fan2017eeg} scaled the range of each subject's features using the means of the subject features across the classes.

\cite{fernandez2021cross} applies single-subject $Z-score$ normalisation after each neural network layer (Stratified Normalisation). 

In \cite{arevalillo2019combining} a simple transformation of the original data for better classification performance is proposed. It uses binary indicator features composed of 0s and 1s, depending on whether the original feature is lower or higher than the median feature value. This leads to a more effective reduction of the subject-dependent part of the EEG signal.

\cite{chen2021ms} the effect of different normalisation strategies is evaluated on DAN and a proposed domain adaptation method (MS-MDA) in an emotion recognition context. The reported results showed that the normalisation scheme could significantly impact the final classification performance.
\section{Notation}
In the remainder of this work we adopt the following notation:
let $X \in \mathbb{R}^{n \times m}$ an EEG dataset having $n$ samples and $m$ features per sample acquired from $N$ subjects, and $X_s \in \mathbb{R}^{n_s \times m}$ a subset of $X$ containing only the $n_s$ samples related to a subject $s$. 
We denote with $\mu$ and $\sigma$ respectively the mean and the standard deviation estimates computed on $X$.
We denote with $\mu_s$ and $\sigma_s$ respectively the mean and the standard deviation estimates computed on $X_s$.

\section{Problem description}
\label{sec:problem_description}
Dealing with the non-stationarity of EEG signals is among the major challenges for the BCI research \cite{blanco1995stationarity,azab2018review,yger2016riemannian}. Non-stationarity of EEG signals over time can be observed in conjunction with changes in behaviours and mental states of the observed subjects. From a statistical point of view, it refers to a continuous change in a class definition causing a change in data distributions \cite{liyanage2013dynamically}, thus implying a high variability of the signals among different experimental sessions for each subject.
Moreover, high signal variability can also be seen among different subjects due to individual differences expressed through EEGs \cite{meltzer2007individual}.

In the context of a classification task for a set of subjects, usually two scenarios are mainly explored: the building of a \textit{subject-independent} model shared by all the subjects, where the goal is to realise a unique model able to be used on any subject with high performance, 
or the fitting of a \textit{subject-dependent} model, where specific models are built for each subject. 
Due to the consequences of the high variability of the EEG signals, in these scenarios the  hypothesis that the training set comes from the same probability distribution of the test set can be violated. 

In the context of ML, due to the problems related to the high variability of the EEG signals, subjects and/or sessions can be considered as belonging to different domains affected by a distribution shift \cite{lan2018domain}. For this reason, in several works regarding EEG signals classification, Domain Adaptation techniques improved classification performances (e.g., \cite{lan2018domain,li2019domain,zhao2020deep}).

In this paper, we aim to investigate the hypothesis that the improvements in classifier performance reported by several affirmed DA methods may be strongly conditioned by data normalisation strategies rather than the DA techniques.
\begin{figure}[!ht]
    \centering
    \scalebox{0.19}{
        \includegraphics{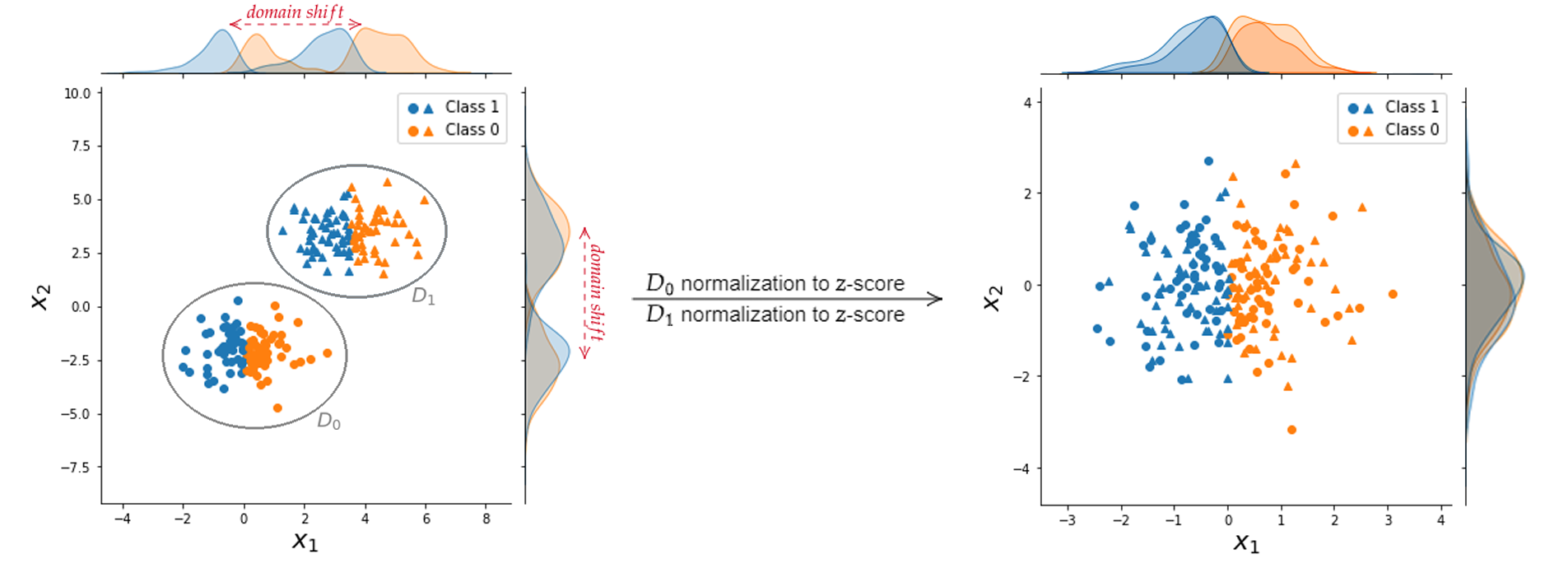}
    }
    \caption{A graphical representation of the hypothesis explored in this work. Given two domains $D_0$ and $D_1$ sharing the same feature and label spaces and affected by a domain shift (left), the application of a $z$-score normalization could reduce the domain shift between the domains (right) regardless any DA technique.}
    \label{fig:intuition}
\end{figure}

For instance, Chen et al. in \cite{chen2021personal} have already highlighted the impact of representing each domain via \textit{z}-score on the classification of the signals, but without analysing its impact on classical DA methods. We remember that the $z$-score $\mathbf{Z}$ of a set of data $X$ can be computed as:
$$\mathbf{Z}(X, \mu, \sigma)= \frac{X-\mu}{\sigma}$$,
where  $\mu$ and $\sigma$ are usually the mean and the standard deviation computed respectively over the features of $X$. In fact, the authors emphasised how the application of the $z$-score to highly clustered domain subjects
could help mitigate individual differences of the signals in the feature space.

In the presence of data coming from several subject domains affected by domain shifts and processed through DA techniques, we wonder whether the data normalization stage might be a critical step when one applies a DA method.
Our idea is intuitively represented in the simple example shown in Figure \ref{fig:intuition}, where we can see two different domains, $D_0$ and $D_1$, having the same feature and label space but affected by a domain shift. 
Assuming that the two domains have the same conditional probabilities $P(Y_{D_0}|X_{D_0}) = P(Y_{D_1}|X_{D_1})$ as in Figure \ref{fig:intuition}, a proper domain $z$-score data normalization stage, a scenario in which conditional distributions mostly overlap could be verified, thus mitigating the domain shift problem without the using of any DA method.

In the remainder of this paper, we investigate the impact of the normalization stage on DA methods through experiments on different EEG datasets. In particular, for each dataset, 
we compare the impact of different normalization strategies applied with and without several DA methods and the performance obtained by the DA strategies as usually described in the literature. 
\section{Experimental assessment}
\label{sec:experimental_assessment}
In this section, we investigate the impact of the normalization stage on DA methods considering the $z$-score as normalization strategy.

In classical ML problems, two main assumptions are that i) we have no access to test data during the training stage, ii) training and test data belong to the same domain. 
In this context, to compute the $z$-score normalization, $\mu$ and $\sigma$ are usually estimated only over the training data due to the assumption that both the training and the test data are samples drawn from the same distribution, therefore sharing the same estimated parameters. 
On the other hand, in an Unsupervised Domain Adaptation scenarios, the training and test set are not usually drawn from the same distributions, and a set of unlabelled test data is supposed to be available during the training.
In this case, $\mu$ and $\sigma$ can be estimated in two ways
:
\begin{enumerate}
    \item $\mu$ and $\sigma$ are estimated separately on training data and unlabelled test data; 
    \item $\mu$ and $\sigma$ are estimated only on training data, as in the ML classical scenarios.
\end{enumerate}
In the context of EEG data, acquisitions are made across several subjects/sessions. Since each subject/session can be considered as a different domain due to non-stationarity of EEG signals, two different hypothesis about the belonging domains can be made:
\begin{enumerate}[a.]
    \item all the subjects/sessions are considered as belonging to the same domain; 
    \item each subject/session is considered as a different domain.
\end{enumerate}
Considering these different conditions, several modalities emerge to perform $z$-score normalization in the contexts of EEG-data and DA methods. The following $z$-score normalization strategies were examined in this paper: 
\begin{itemize}
    \item $\mathbf{Z_0}$: the training set was transformed computing $\mu$ and $\sigma$ on the only training data; the test subject/session was transformed using parameters $\mu$ and $\sigma$ computed over the training set (i.e., it corresponds to the the classical $z$-score normalization applied on the training data);
    \item $\mathbf{Z_1}$: each subject/session $s$ belonging to the training set was transformed using its own parameters $\mu_s$ and $\sigma_s$; the test subject/session was transformed using $\mu$ and $\sigma$ computed on the whole training data;
    \item $\mathbf{Z_2}$: each subject/session $s$, regardless the training/test set partitioning, was transformed using its own parameters $\mu_s$ and $\sigma_s$;
    \item $\mathbf{Z_3}$: the training set was transformed using parameters $\mu$ and $\sigma$ computed on the whole training data; the test subject/session was transformed on its own parameters $\mu_s$ and $\sigma_s$.
\end{itemize}
Our hypothesis was explored in a series of experiments on three EEG datasets: SEED \cite{zheng2015investigating}, BCI Competition IV 2a \cite{brunner2008bci} and DEAP \cite{koelstra2011deap}. Further details regarding the mentioned datasets are provided in this section.
We point out that our interest in these experiments is in investigating the normalisation strategies' impact on DA methods in terms of performance degradation/enhancement of classifiers and not in providing new state-of-the-art results on the involved datasets.

\begin{figure}[!ht]
    \centering
    \scalebox{0.5}{
        \includegraphics{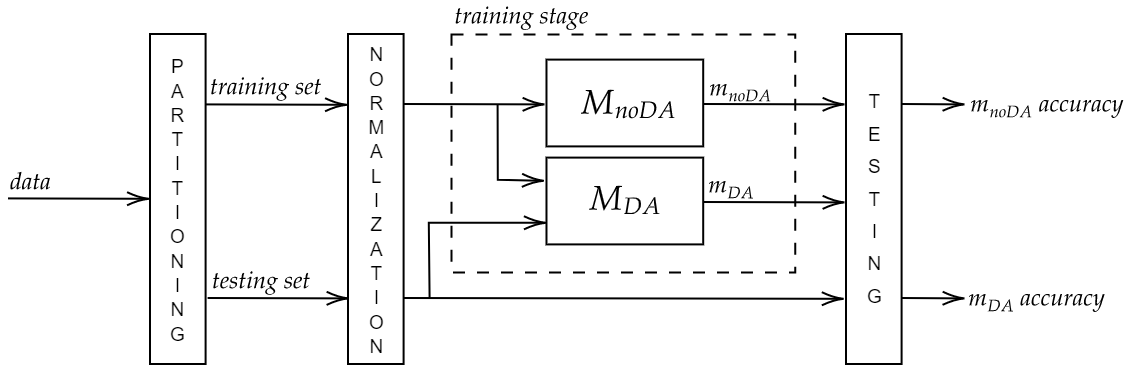}
    }
    \caption{Graphical representation of the processing pipeline adopted in this work. After the data partitioning and normalization stages, the impact of the data normalization is inspected on a given ML technique $M$ with and without DA methods (respectively, $M_{DA}$ and $M_{noDA}$). Then, performances are evaluated on the testing set through the models $m_{DA}$ and $m_{noDA}$ fitted during the training stages and compared.}
    \label{fig:pipeline}
\end{figure}

For each dataset, we conducted our experiments on the four normalisation strategies described above using different frameworks typically used in DA: i) a deepDA-based framework, where we analysed the performances of the two well-known deep DA methods DANN \cite{ganin2016domain} and ADDA \cite{tzeng2017adversarial}, applied on ANNs, and comparing their performance with the one achieved using the same ANN architectures without the DA components, ii) a shallow DA-based framework, where we compared the performances obtained using a typical projection-based method as TCA \cite{pan2010domain} and KPCA \cite{scholkopf1998nonlinear}, followed by a Support Vector Machine (SVM) \cite{noble2006support} classifier, with those achieved using the SVM classifier only. Figure \ref{fig:pipeline} shows the general processing pipeline adopted in this work.

Model performances were obtained adopting i) the \textit{Leave-One-Subject-Out Cross-Validation} (LOSO-CV) strategy for the subject-independent case where, for each iteration, the training set resulted to be a composition of multiple training subjects while the test set was composed by just one test subject, ii) the \textit{Hold-Last-Session-Out} (HLSO) strategy for the subject-dependent case, where the last session from a chronological point of view was considered as test set while the others are considered as training set. These experiments were not performed on the DEAP dataset since just one session is provided.

For the shallow DA-based experiments, we followed the setup proposed in \cite{zheng2015transfer}, searching for the best kernel methods among \{Linear, RBF, Gaussian\}, 
while for ANNs-based ones, according to the original architectures of the DANN and ADDA methods, \textit{full-connected multi-layered neural networks} were chosen as models for each architectural component (feature extractor, label predictor, domain classifier). Hyperparameters were tuned using a bayesian optimisation method \cite{snoek2012practical}. In particular, for each architectural component, the number of layers was constrained to a maximum of $3$, the number of nodes per layer was searched in the set $\{1,2,...,1000\}$ and the activation function was searched among ReLU, Sigmoid and LeakyReLU. Moreover, for the ANNs-based experiments, each experiment was made considering early stopping  as convergence criterion with $20$ epochs as patience; the $10$ \% of the training set was extracted and considered as validation set using stratified sampling \cite{parsons2014stratified} on class labels; optimisation was performed using Adam optimiser \cite{kingma2014adam} with a learning rate that was searched in the space $\{0.1, 0.01, ..., 0.0001\}$. In order to ensure fairness in experimental conditions, the best architecture found in the DANN method was also used in ADDA and in the pure ANN without DA components (i.e. domain classifier).
The accuracy score was used for each experiment to evaluate the performance of the method.

\subsection{SEED}
The SEED dataset consists of EEG data from 15 subjects while watching 15 video clips of about 4 minutes. Each video clip was chosen to induce positive, neutral and negative emotions. For each subject, three data sessions were collected with an interval of about one week. EEG signals were recorded in 62 channels using the ESI Neuroscan System\footnote{https://compumedicsneuroscan.com/} 
according to the international 10-20 system, at a sampling rate of 1000 Hz and downsampled to 200 Hz. 
Following \cite{ma2019reducing}, we considered the pre-computed Differential Entropy (DE) features smoothed by Linear Dynamic Systems (LDS). DE features are pre-computed, for each second, in each channel, over the following five bands: Delta (1–3 Hz); Theta (4–7 Hz); Alpha (8–13 Hz); Beta (14–30 Hz); Gamma (31–50 Hz). As in \cite{ma2019reducing}, following a sampling stratified on class labels, 1000 samples for each subject were randomly selected as training set due to the limited available memory and computation time.

\subsection{BCI Competition IV 2a} 
The BCI Competition IV 2a dataset consists of EEG data acquired from 9 subjects during motor imagery tasks. The dataset involves 4 EEG measurement classes: left hand, right hand, feet, and tongue. 
22 Ag/AgCl electrodes recorded EEG signals at a sampling rate of 250 Hz.
The EEG signals were filtered using the IIR Butterworth filter of order 5 with a bandpass cut-off frequency of 8 to 30 Hz.
The four-class classification problem was reduced to a binary classification problem, thus considering left-hand and right-hand labels as in, for example, \cite{selim2016reducing,malan2021time,mishuhina2021complex}.
Finally, the Common Spatial Pattern (CSP) \cite{blankertz2007optimizing} was applied since it is a widely recognised feature extraction technique involved in classification tasks on the motor imagery studies \cite{padfield2019eeg}.

\subsection{DEAP}
The DEAP dataset consists of EEG data acquired from 32 subjects while they were exposed to 40 of about 1 minute. 
EEG signals were recorded in 32 channels using the Biosemi ActiveTwo devices\footnote{https://www.biosemi.com} at a sampling rate of 512 Hz and downsampled to 128 Hz.
After watching each video, each subject was
required to rate each video in terms of valence (pleasantness level), arousal (excitation level), dominance (control power), liking (preference) and familiarity (knowledge of the stimulus), where each  rating ranged from one (weakest) to nine (strongest). Only the familiarity level ranged from one to five. The EEG signals were recorded by Biosemi ActiveTwo
devices at a sampling rate of 512 Hz and downsampled to 128 Hz.
\newline
Following \cite{lan2018domain}, we labelled each trial discretizing and partitioning the dimensional emotion space as follows:
\begin{itemize}
    \item positive if valence rating is greater than 7;
    \item neutral if valence rating is smaller than 7 and greater than 3;
    \item negative if valence rating is smaller than 3.
\end{itemize}
Moreover, as in \cite{lan2018domain}, since trials 18, 16 and 38 had the most participants reporting to have successfully induced positive, neutral and negative emotion, a subset of subjects that reported a successful emotion induction with these trials were was selected. In particular, data related to subjects 2, 5, 10, 11, 12, 13, 14, 15, 19, 22, 24, 26, 28, and 31 were involved in the experimental assessments.
Finally, DE was applied to EEG data in the bands Delta, Theta, Alpha, Beta and Gamma, as for the SEED dataset.
\section{Results}
\label{sec:results}
In this Section, we present the results collected in our series of experiments. For each experiment, 
the results obtained without normalization are reported under the heading of \textit{"noNorm"}.
For the ANNs and SVMs based experiment, with \textit{"noDA-ANN"} and \textit{"noDA-SVM"} we refer to the pure architectures without the DA components (thus, only with their feature-extractor and label predictor).
For the subject-independent experiments, we report the mean and the standard deviation over the folds for each type of normalisation.
On the other hand, for the subject-dependent experiments, we report the mean and the standard deviation over the subjects for each kind of normalisation.

\subsection{SEED}
In Table \ref{tab:seed_loso_results} the results related to the subject-independent experiments on SEED are reported. 
Regarding the deep-DA experiments, for the $Z_0$ and $Z_2$ normalisations, the use of the ANN leads to better results than those obtained through the DANN and ADDA methods. For the $Z_3$ normalizaion the use of the ANN leads to results comparable with the ones reached by using DANN, but higher than those reached by using ADDA; for the $Z_1$ normalisation, NoDA-ANN performances are lower than those of DANN, but higher than the ones achieved using ADDA (the same situation is also encountered for the $noNorm$ case).
The best performances are attributed to the NoDA-ANN case on the $Z_2$ normalisation with a mean accuracy of $81.52 \pm 7.26$. Thus, the use of only $Z_2$ normalisation outperforms the other tested methods.
\newline
For the shallow-DA based experiments, instead, we can observe that for the $Z_0$ normalisation, the use of SVM leads to lower results than using TCA-SVM, but higher than using KPCA-SVM; for the $Z_1$ and $Z_2$ normalisations, NoDA-SVM leads to lower results than both DA techniques; for the $Z_3$ normalisation, NoDA-SVM reaches lower results than TCA-SVM, but comparable with KPCA-SVM; for the $noNorm$ case, NoDA-SVM leads results higher than TCA-SVM but lower than KPCA-SVM. 
The best performances are attributed to the KPCA-SVM case on the $Z_2$ normalisation with a mean accuracy of $80.74 \pm 6.11$. However, the most significant improvement seems to be obtained by the $Z_2$ normalisation in the NoDA-SVM, improving the performance to $74.71\pm 8.47\ \%$ from an initial $52.96\pm 9.82\ \%$ accuracy without any normalisation, while the use of DA methods gives an improvement of about $6 \ \%$.
\begin{table}[!h]
\centering
\setlength{\tabcolsep}{5pt}
\captionsetup{font=scriptsize}
\caption{SEED - Leave-One-Subject-Out Cross-Validation Accuracy, Mean \% (Std \%)}
\scalebox{0.82}{
    \begin{tabular}{clll|lll}
    \bottomrule
    & \multicolumn{3}{c}{deep DA} & \multicolumn{3}{c}{shallow DA}\\
    \cline{2-7}
    & noDA-ANN & DANN & ADDA & noDA-SVM & TCA-SVM & KPCA-SVM\\
    \hline
    noNorm & 45.50 (13.18) & 50.65 (12.19) & 33.13 (0.22) & 52.96 (9.82) & 46.68 (13.34) & 58.61 (7.50)\\
    $Z_0$ & 48.31 (14.09) & 43.60 (11.86) & 43.97 (12.86) & 52.47 (11.33) & 71.58 (7.16) & 48.16 (13.22)\\
    $Z_1$ & 50.54 (15.13) & 60.35 (21.45) & 46.53 (12.92) & 52.32 (15.06) & 79.70 (8.98) & 53.59 (16.91)\\
    $Z_2$ & 81.52 (7.26) & 79.03 (7.71) & 70.43 (14.17) & 74.71 (8.47) & 80.09 (6.51) & 80.74 (6.11)\\
    $Z_3$ & 75.22 (7.85) & 75.79 (4.78) & 60.57 (13.92) & 73.24 (8.37) & 76.37 (7.44) & 73.91 (7.31)\\
    \bottomrule
    \end{tabular}
    \label{tab:seed_loso_results}
    }
\end{table}
\newline
In Table \ref{tab:seed_cs_results} the results related to the subject-dependent experiments on SEED are reported. 
For the deep-DA experiments, on the $noNorm$, $Z_0$, $Z_1$ and $Z_2$ cases NoDA-ANN leads to higher results than DA method; for the $Z_3$ normalisation, the use of ANN leads to lower results than the DANN method, but higher than the ADDA method.
The best performances are attributed to the NoDA-ANN case on the $Z_2$ normalisation with a mean accuracy of $83.93 \pm 9.60$.
\newline
Regarding the shallow-DA based experiments, on the $noNorm$, $Z_2$ and $Z_3$ normalisations noDA-SVM achieves better results than SVM applied with DA methods; for the $Z_0$ normalisation, NoDA-SVM performances are lower than those of TCA-SVM but higher than the ones of KPCA-SVM; for the $Z_0$ normalisation, the use of NoDA-SVM leads to results lower than those of TCA-SVM but comparable with the ones obtained through KPCA-SVM.
The best performances are attributed to the NoDA-SVM case on the $Z_3$ normalisation with a mean accuracy of $86.56 \pm 8.15$. Therefore, also in this case the use of a simple normalisation method seems to be more effective than the selected DA-methods.
\begin{table}[!h]
\centering
\setlength{\tabcolsep}{5pt}
\captionsetup{font=scriptsize}
\caption{SEED - Cross-Session Accuracy, Mean \% (Std \%)}
\scalebox{0.82}{
    \begin{tabular}{clll|lll}
    \bottomrule
    & \multicolumn{3}{c}{Deep DA} & \multicolumn{3}{c}{Shallow DA}\\
    \cline{2-7}
    & noDA-ANN & DANN & ADDA & SVM & TCA-SVM & KPCA-SVM\\
    \hline
    noNorm & 45.47 (20.43) & 43.15 (17.43) & 37.10 (10.83) & 63.63 (18.92) & 47.52 (12.45) & 61.31 (14.33)\\
    $Z_0$ & 64.85 (17.38) & 51.76 (17.71) & 57.33 (19.52) & 62.32 (17.33) & 79.56 (10.21) & 61.66 (16.88)\\
    $Z_1$ & 66.37 (16.16) &  55.25 (19.95) &  65.07 (17.50) &  63.39 (20.02) &  76.96 (12.99) &  63.25 (15.92)\\
    $Z_2$ & 83.93 (9.60) & 83.84 (10.55) & 76.80 (12.50) & 85.59 (9.89) & 81.67 (11.83) & 83.62 (9.45)\\
    $Z_3$ & 83.09 (9.98) & 84.43 (9.67) & 77.80 (12.82) & 86.56 (8.15) & 83.15 (10.92) & 84.09 (10.45)\\
    \bottomrule
    \end{tabular}
    \label{tab:seed_cs_results}
    }
\end{table}

\subsection{BCI Competition IV 2a}
Differently from experiments on SEED, only results related to $Z_1$, $Z_2$ and $Z_3$ normalizations are reported since the CSP implementation\footnote{In this work we performed CSP on data using the implementation provided by the Python package MNE \cite{GramfortEtAl2013a}} we already performed a \textit{z}-score normalization, thus making the $Z_0$ normalization unnecessary in our experiments. 
In Table \ref{tab:bci_loso_results} the results related to the subject-independent experiments on BCI Competition IV 2a are reported.
\begin{table}[!h]
\centering
\setlength{\tabcolsep}{5pt}
\captionsetup{font=scriptsize}
\caption{BCI Comp. IV 2a - Leave-One-Subject-Out Cross-Validation Accuracy, Mean \% (Std \%)}
\scalebox{0.82}{
    \begin{tabular}{clll|lll}
    \bottomrule
    & \multicolumn{3}{c}{Deep DA} & \multicolumn{3}{c}{Shallow DA}\\
    \cline{2-7}
    & noDA-ANN & DANN & ADDA & noDA-SVM & TCA-SVM & KPCA-SVM\\
    \hline
    noNorm & 61.42 (13.36) & 62.19 (13.43) & 63.22 (12.90) & 61.03 (12.70) & 56.10 (10.62) & 61.11 (12.62)\\
    $Z_1$ & 62.11 (13.54) & 61.73 (13.02) & 62.84 (13.77) & 61.72 (10.30) & 61.03 (11.76) & 61.73 (10.93)\\
    $Z_2$ & 67.98 (11.81) & 67.52 (11.40) & 68.58 (10.96) & 68.52 (11.35) & 67.90 (11.89) & 68.44 (12.02)\\
    $Z_3$ & 63.36 (11.48) & 68.13 (13.08) & 59.77 (12.80) & 67.90 (12.35) & 67.44 (13.55) & 67.82 (12.48)\\
    \bottomrule
    \end{tabular}
    \label{tab:bci_loso_results}
    }
\end{table}
\newline
For the deep-DA experiments, NoDA-ANN always leads to lower or comparable results with DA methods, except for the $Z_3$ normalization where its performance are lower than DANN but higher than ADDA. 
The best performances are attributed to the DANN case on the $Z_3$ normalization with a mean accuracy of $68.13 \pm 13.08$. However, the use of the only DANN method without any normalisation gives an improvement less than $1 \ \% $, while the use of the only normalisation can lead an improvement of about $6 \ \%$, showing that the normalisation can have a significant effect on the final performance.
\newline
For the shallow-DA based experiments instead, for each normalization type including $noNorm$, NoDA-SVM always reaches results higher or comparable with DA methods.
The best performances are attributed to the NoDA-SVM case on the $Z_2$ normalization with a mean accuracy of $68.52 \pm 11.35$.
\begin{table}[!h]
\centering
\setlength{\tabcolsep}{5pt}
\captionsetup{font=scriptsize}
\caption{BCI Comp. IV 2a - Cross-Session Accuracy, Mean \% (Std \%)}
\scalebox{0.82}{
    \begin{tabular}{clll|lll}
    \bottomrule
    & \multicolumn{3}{c}{deep DA} & \multicolumn{3}{c}{shallow DA}\\
    \cline{2-7}
    & noDA-ANN & DANN & ADDA & noDA-SVM & TCA-SVM & KPCA-SVM\\
    \hline
    noNorm & 57.56 (10.47) & 55.94 (8.87) & 54.79 (15.46) & 55.79 (9.38) & 50.54 (1.52) & 60.88 (13.31)\\
    $Z_1$ & 54.63 (10.32) & 56.94 (8.63) & 59.39 (8.87) & 55.48 (9.62) & 55.79 (9.63) & 61.50 (10.04)\\
    $Z_2$ & 64.97 (14.28) & 64.12 (15.39) & 58.24 (15.71) & 63.43 (15.10) & 63.66 (15.17) & 68.36 (11.47)\\
    $Z_3$ & 63.27 (13.77) & 62.27 (13.96) & 61.30 (16.64) & 67.82 (12.48) & 67.43 (13.55) & 68.13 (12.22)\\
    \bottomrule
    \end{tabular}
    \label{tab:bci_cs_results}
    }
\end{table}
\newline
In Table \ref{tab:bci_cs_results} the results related to the subject-dependent experiments on BCI Competition IV 2a are reported.
Regarding the deep-DA experiments, in the $noNorm$, $Z_2$ and $Z_3$ cases ANN leads to higher results than DA methods; for the $Z_1$ normalization using NoDA-ANN lower performances than both of the DA methods are achieved.
The best performances are attributed to the NoDA-ANN case on the $Z_2$ normalization with a mean accuracy of $64.97 \pm 14.28$. 
\newline
For the shallow-DA based experiments, NoDA-SVM always reach accuracies lower than DA methods, except on the $noNorm$ and $Z_3$ cases where it leads to higher accuracies than TCA-SVM.
The best performances is attributed to the KPCA-SVM case on the $Z_2$ normalization with a mean accuracy of $68.36 \pm 11.47$.

\subsection{DEAP}
Differently from SEED and BCI Competition IV 2a, experiments on DEAP were performed only for the subject-independent case since the dataset provided a single session. The results are reported in Table \ref{tab:deap_loso_results}.
\begin{table}[!h]
\centering
\setlength{\tabcolsep}{5pt}
\captionsetup{font=scriptsize}
\caption{DEAP - Leave-One-Subject-Out Cross-Validation Accuracy, Mean \% (Std \%)}
\scalebox{0.82}{
    \begin{tabular}{clll|lll}
    \bottomrule
    & \multicolumn{3}{c}{deep DA} & \multicolumn{3}{c}{shallow DA}\\
    \cline{2-7}
    & noDA-ANN & DANN & ADDA & noDA-SVM & TCA-SVM & KPCA-SVM\\
    \hline
    noNorm & 34.21 (4.11) & 34.21 (3.15) & 33.93 (2.15) & 31.31 (9.76) & 36.90 (12.34) & 41.23 (10.02)\\
    $Z_0$ & 30.44 (10.13) & 31.11 (13.38) & 34.92 (13.64) & 32.56 (1.94) & 41.23 (13.26) & 38.33 (9.86)\\
    $Z_1$ & 35.60 (8.72) & 36.51 (7.90) & 34.92 (13.64) & 32.55 (10.83) & 42.66 (11.70) & 34.52 (7.42)\\
    $Z_2$ & 36.67 (12.45) & 35.52 (13.84) & 32.94 (11.59) & 32.13 (14.77) & 42.46 (15.99) & 40.67 (15.13)\\
    $Z_3$ & 39.33 (14.08) & 41.27 (14.91) & 38.89 (14.16) & 33.73 (14.75) & 42.62 (17.06) & 43.77 (12.68)\\
    \bottomrule
    \end{tabular}
    \label{tab:deap_loso_results}
    }
\end{table}
\newline
For the deep-DA experiments, on the $Z_0$ normalization, NoDA-ANN shows lower results than the DA methods; for the $Z_1$ and $Z_3$ normalizations, ANN performances are lower than those of DANN but higher than the ones obtained through ADDA; for the $Z_2$ normalization, NoDA-ANN results are higher than those of DA methods; for the $noNorm$, NoDA-ANN results are equal to DANN ones and higher than ADDA ones.
The best performances are attributed to the DANN case on the $Z_3$  normalization with a mean accuracy of $41.27 \pm 14.91$.
\newline
Regarding the shallow-DA based experiments, noDA-SVM always leads to lower results than DA methods.
The best performances are attributed to the KPCA-SVM case on the $Z_3$ normalization with a mean accuracy of $43.77 \pm 12.68$. In this case the DA methods give an important improvement in the performance, further increased by the normalisation methods, showing the importance of using both of them.
\section{Discussion}
\label{sec:discussion}
The experimental results suggest that the normalisation method one uses plays a crucial role in improving the classification performances in DA approaches on EEG data.
We will focus our discussions on results related to the SEED dataset, but similar observations can also be made for BCI Competition IV 2a and DEAP. 

\subsection{Subject-independent experiments}
In Table \ref{tab:normalization_visualization}, data related to a fold of a subject-independent experiment are represented using t-SNE \cite{van2008visualizing} before the application of any DA method for each normalization type.
\newline
\begin{table}
    \setlength{\tabcolsep}{2pt}
    \scalebox{0.83}{
        \begin{tabular}{c|cccc}
            & $Z_0$ & $Z_1$ & $Z_2$ & $Z_3$\\ 
            \hline
            Domains & 
            \includegraphics[trim={0.8cm 0.8cm 0.8cm 0.8cm},clip,width=.25\linewidth,valign=m]{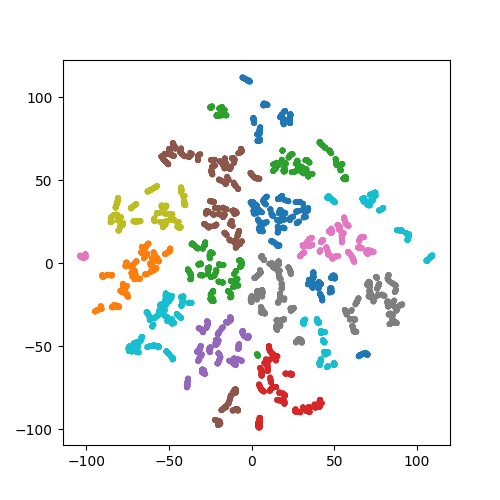} & \includegraphics[trim={0.8cm 0.8cm 0.8cm 0.8cm},clip,width=.25\linewidth,valign=m]{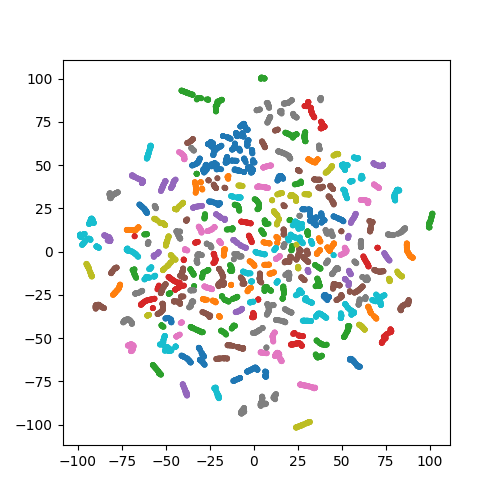} &
            \includegraphics[trim={0.8cm 0.8cm 0.8cm 0.8cm},clip,width=.25\linewidth,valign=m]{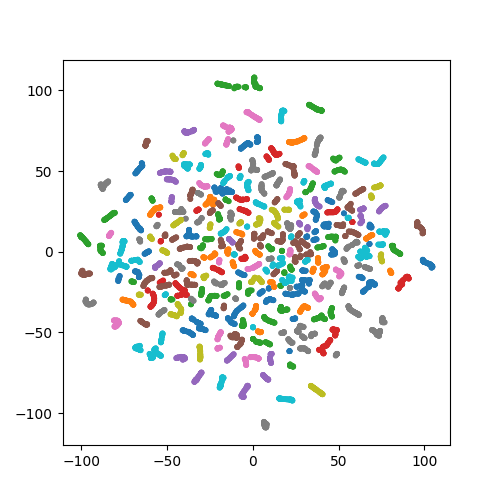} & \includegraphics[trim={0.8cm 0.8cm 0.8cm 0.8cm},clip,width=.25\linewidth,valign=m]{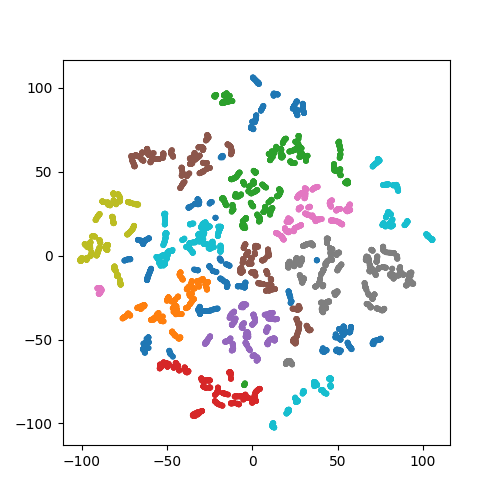}\\
            Training/Test &
            \includegraphics[trim={0.8cm 0.8cm 0.8cm 0.8cm},clip,width=.25\linewidth,valign=m]{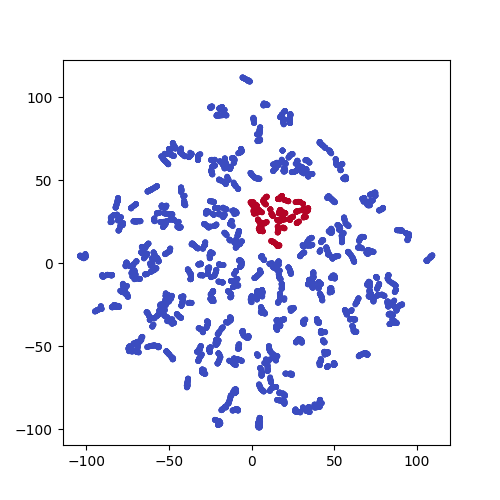} & \includegraphics[trim={0.8cm 0.8cm 0.8cm 0.8cm},clip,width=.25\linewidth,valign=m]{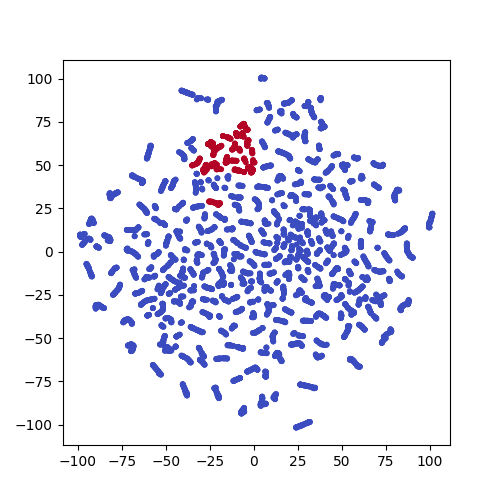} &
            \includegraphics[trim={0.8cm 0.8cm 0.8cm 0.8cm},clip,width=.25\linewidth,valign=m]{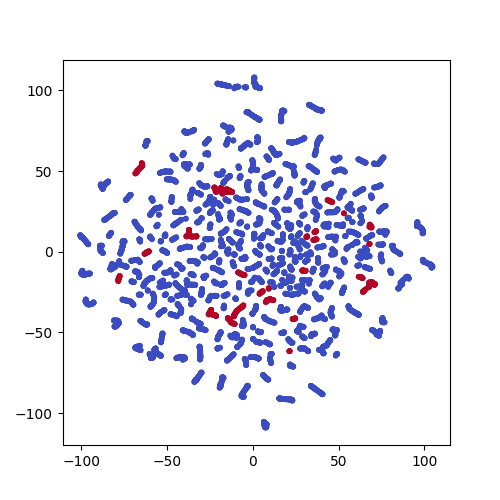} & \includegraphics[trim={0.8cm 0.8cm 0.8cm 0.8cm},clip,width=.25\linewidth,valign=m]{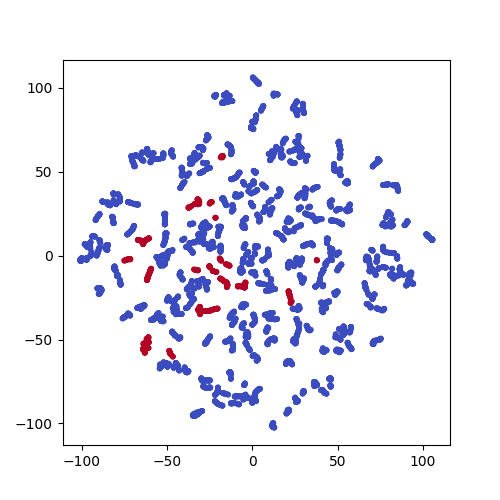}\\
            Labels &
            \includegraphics[trim={0.8cm 0.8cm 0.8cm 0.8cm},clip,width=.25\linewidth,valign=m]{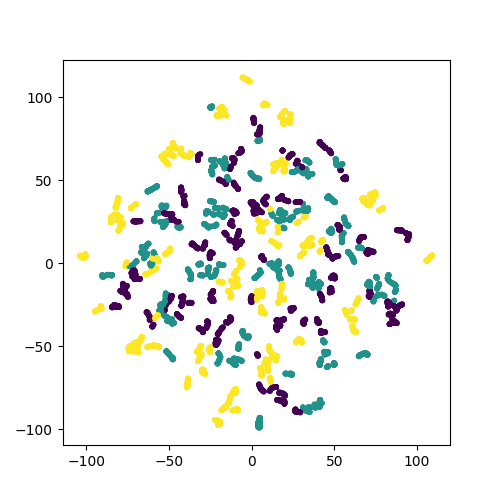} & \includegraphics[trim={0.8cm 0.8cm 0.8cm 0.8cm},clip,width=.25\linewidth,valign=m]{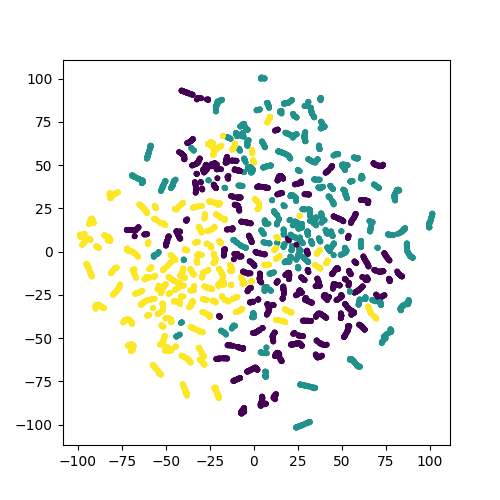} &
            \includegraphics[trim={0.8cm 0.8cm 0.8cm 0.8cm},clip,width=.25\linewidth,valign=m]{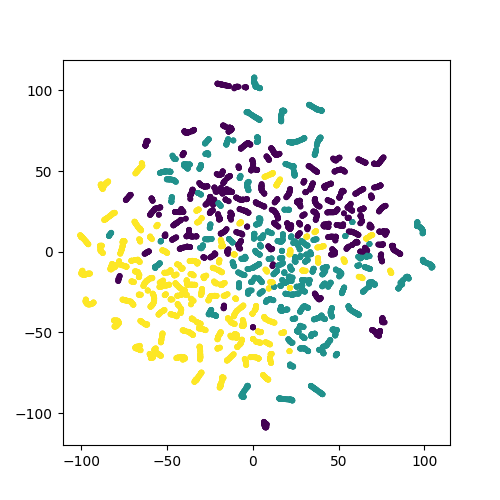} & \includegraphics[trim={0.8cm 0.8cm 0.8cm 0.8cm},clip,width=.25\linewidth,valign=m]{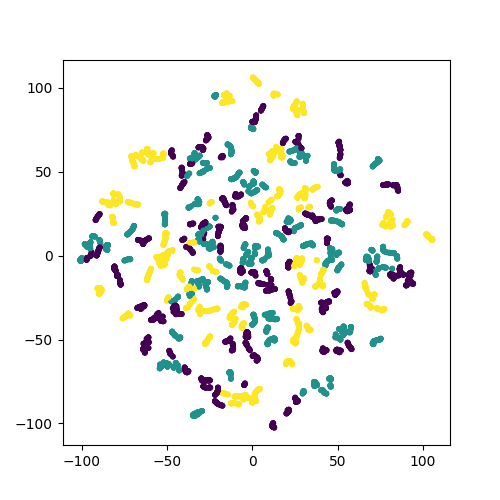}\\
        \end{tabular}    }
    
    \caption{Graphical representation of SEED data on subject-independent experiments after each type of normalization. On the first row, all domains involved in the series of experimental are marked by different colors; on the second row, training and test data are marked with blue and red, respectively; on the third row, data are distinguished by their label. On each column, data transformed by the relative normalization type are shown.}
    \label{tab:normalization_visualization}
\end{table}
It is interesting to notice how for $Z_1$ and $Z_2$ normalisations, a similar scenario to Figure \ref{fig:intuition} is verified on these data: after the normalisation stage, clusters of data having the same labels are observable, corroborating the hypothesis that conditional distributions over the subjects could be equal or similar, thus leading the normalisation to reduce the domain shifts without DA methods.

In the ANNs based experiments, we can notice that ANN model achieves the best performances on $Z_2$ normalisation without using any DA method.
Moreover, this can also be observed from how performances are distributed on the ANN method as the type of normalisation changes: accuracy means are distributed from a minimum of $\sim 46\ \%$ to a maximum of $\sim 82\ \%$.

On the other hand, for the Projection Matrix-based experiments, the highest performances are reached by the KPCA-SVM method on $Z_2$ normalisation. According to how accuracy means vary as the normalisation type changes (from a minimum of $\sim 59\ \%$ to a maximum of $\sim 81\ \%$), we can hypothesise that also the right balance between DA methods and normalisation type has an impact on performances.

Comparing the ANNs based experiments with the Projection Matrix-based ones, we can conclude that the impact DA methods could be affected by the choice of the model: in the first case, using ANNs, the DA methods does not give any contribution; in the second case, the DA method contributes to improving the SVM performances.

\subsection{Subject-dependent experiments}
In Table \ref{tab:intra_normalization_visualization}, data related to a subject sampled during the subject-dependent experiments are represented before applying any DA method for each normalisation type.
\newline
\begin{table}
    \setlength{\tabcolsep}{2pt}
    \scalebox{0.83}{
        \begin{tabular}{c|cccc}
            & $Z_0$ & $Z_1$ & $Z_2$ & $Z_3$\\ 
            \hline
            Domains & 
            \includegraphics[trim={0.8cm 0.8cm 0.8cm 0.8cm},clip,width=.25\linewidth,valign=m]{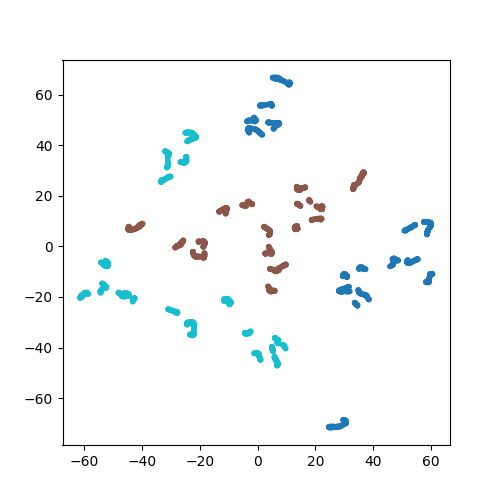} & \includegraphics[trim={0.8cm 0.8cm 0.8cm 0.8cm},clip,width=.25\linewidth,valign=m]{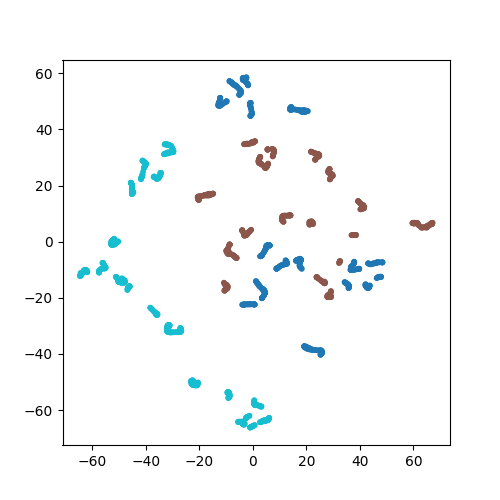} &
            \includegraphics[trim={0.8cm 0.8cm 0.8cm 0.8cm},clip,width=.25\linewidth,valign=m]{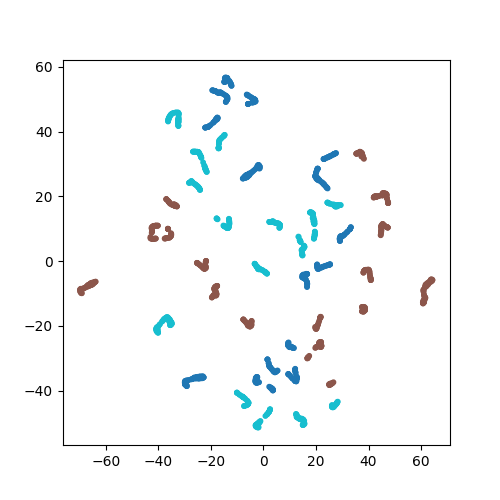} & \includegraphics[trim={0.8cm 0.8cm 0.8cm 0.8cm},clip,width=.25\linewidth,valign=m]{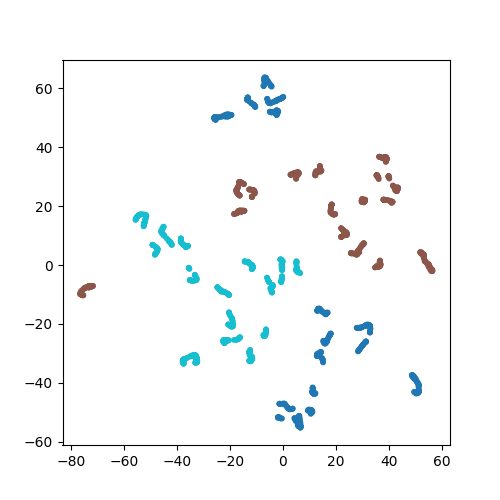}\\
            Training/Test &
            \includegraphics[trim={0.8cm 0.8cm 0.8cm 0.8cm},clip,width=.25\linewidth,valign=m]{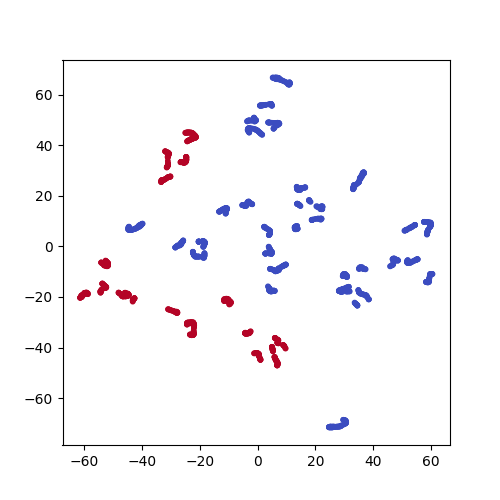} & \includegraphics[trim={0.8cm 0.8cm 0.8cm 0.8cm},clip,width=.25\linewidth,valign=m]{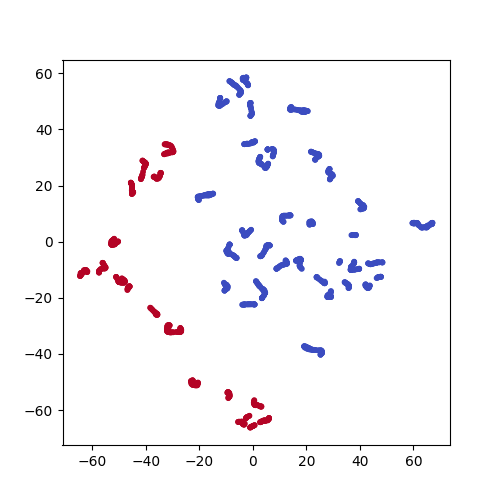} &
            \includegraphics[trim={0.8cm 0.8cm 0.8cm 0.8cm},clip,width=.25\linewidth,valign=m]{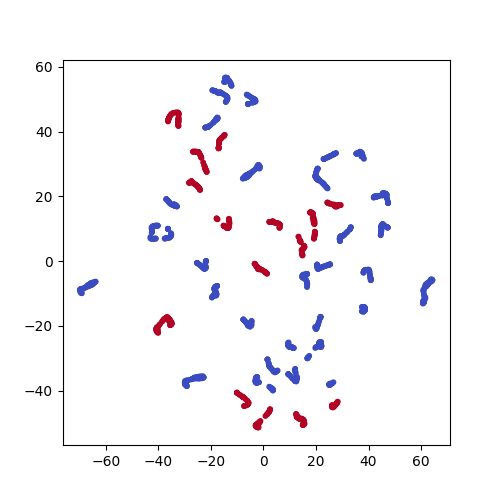} & \includegraphics[trim={0.8cm 0.8cm 0.8cm 0.8cm},clip,width=.25\linewidth,valign=m]{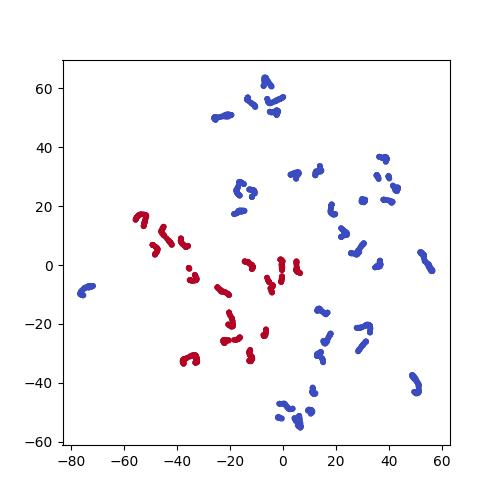}\\
            Labels &
            \includegraphics[trim={0.8cm 0.8cm 0.8cm 0.8cm},clip,width=.25\linewidth,valign=m]{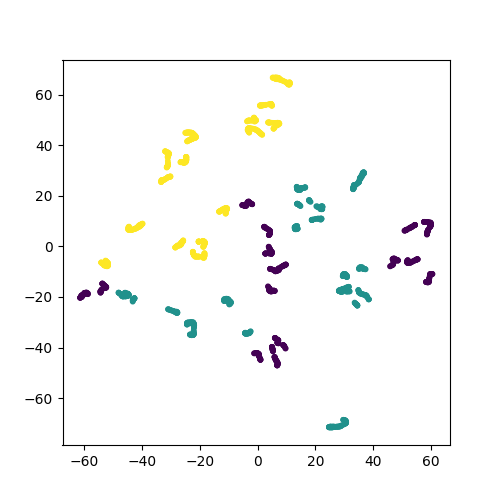} & \includegraphics[trim={0.8cm 0.8cm 0.8cm 0.8cm},clip,width=.25\linewidth,valign=m]{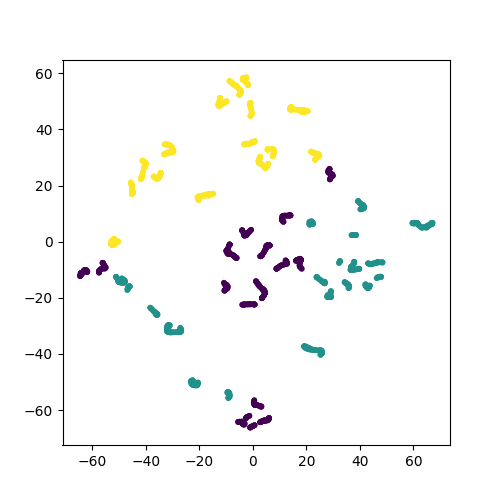} &
            \includegraphics[trim={0.8cm 0.8cm 0.8cm 0.8cm},clip,width=.25\linewidth,valign=m]{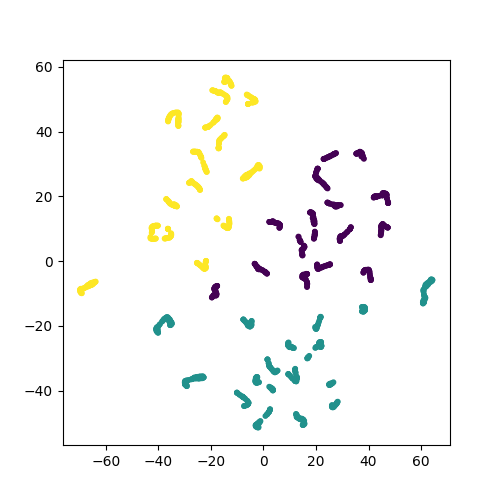} & \includegraphics[trim={0.8cm 0.8cm 0.8cm 0.8cm},clip,width=.25\linewidth,valign=m]{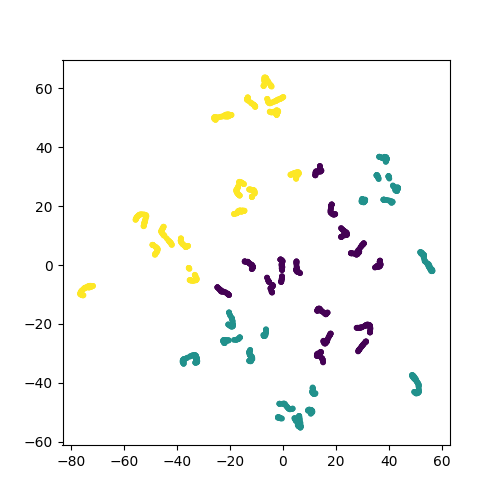}\\
        \end{tabular}    }
    
    \caption{Graphical representation of SEED data on subject-dependent experiments after each type of normalization. On the first row, all domains involved in the series of experimental are marked by different colors; on the second row, training and test data are marked with blue and red, respectively; on the third row, data are distinguished by their label. On each column, data transformed by the relative normalization type are shown.}
    \label{tab:intra_normalization_visualization}
\end{table}
Also in this case, a scenario similar to Figure \ref{fig:intuition} is verified, particularly on $Z_2$ normalisation where data having the same labels are clustered, thus leading to suppose that domains could have equal or similar conditional distributions.

In this case, on deep DA-based experiments, the best results are achieved by noDA-ANN on $Z_2$ normalisation without using DA methods. In contrast, in the shallow DA-based experiments, best results are achieved by noDA-SVM on $Z_3$ normalisation. Also, for subject-dependent experiments, for the best methods, performances change as the normalisation type changes: accuracy means vary from a minimum of $\sim 45\ \%$ to a maximum of $\sim 84\ \%$ (Artificial Neural Network based) and from a minimum of $\sim 64\ \%$ to a maximum of $\sim 87\ \%$ (shallow DA based). Thus, as in the subject-independent case, we can observe that the normalisation type significantly impacts the classifier performances in DA problems. Consequently, a careful choice of normalisation type, DA method and classification model should be made.
To sum up, we can state that when one develops and tests a DA method to classify EEG data, the effect of the normalisation step on the classification performances should be carefully weighed, and a suitable choice of the normalisation method could drastically improve the effectiveness of the DA method or, even avoid the use of DA methods.
\section{Conclusions}
\label{sec:conclusions}
In this work, we examined the effect of data normalisation in several DA approaches.
Starting from the hypothesis that the prior data normalisation could strongly condition the performances reported by several DA methods, considering the $z$-score as the base normalisation procedure, we firstly defined four $z$-score variations. Then we conducted several experiments on different EEG datasets to analyse the effect of each normalisation strategy applied with and without DA methods.
In particular, we dealt with two scenarios typically encountered in EEG classification problems, the subject-independent and subject-dependent cases, where each subject and session can be considered as a different domain due to the non-stationarity of EEG signals.

The results show that the normalisation stage highly impacts classifier performances in several DA scenarios.
The best results are achieved by pure ANNs (deep DA) and SVMs (shallow DA) in several cases, combined with an appropriate normalisation schema, without the need for the investigated DA techniques. 
However, in other cases, the best results are achieved by DA methods combined with a particular type of normalisation, allowing us to consider that searching for the right balance between DA methods and normalisation type could improve classifier performances.

Understanding the impact that the normalisation strategies have on DA approaches could be helpful to improve the performances obtained through DA methods or, in some cases, to avoid DA methods that often turn out to be highly time and hardware-consuming and leading, moreover, to simpler models.

\section*{Data availability}
The datasets used during the current study are available at:
\begin{itemize}
    \item SEED: \url{https://bcmi.sjtu.edu.cn/home/seed/}
    \item BCI Competition IV 2a: \url{https://www.bbci.de/competition/iv/}
    \item DEAP: \url{https://www.eecs.qmul.ac.uk/mmv/datasets/deap/}
\end{itemize}
\section*{Acknowledgments}
This work  is supported by the European Union - FSE-REACT-EU, PON Research and Innovation 2014-2020 DM1062/2021 contract number 18-I-15350-2 and by the Ministry of University and Research, PRIN research project "BRIO – BIAS, RISK, OPACITY in AI: design, verification and development of Trustworthy AI.", Project no. 2020SSKZ7R .

%
%
%
\bibliographystyle{unsrt}
\bibliography{biblio}
\end{document}